\pdfoutput=1

\documentclass[11pt]{article}

\usepackage[usenames,dvipsnames]{xcolor}
\usepackage[]{acl}

\usepackage{times}
\usepackage{latexsym}
\usepackage{hyperref}
\usepackage{url}
\usepackage{microtype}
\usepackage{hyperref}
\usepackage{xspace}
\usepackage{amsmath}
\usepackage{amsthm}
\usepackage{amssymb}
\usepackage{booktabs}
\usepackage{multirow}
\usepackage{multicol}
\usepackage{ifthen}
\usepackage{graphicx}
\usepackage{multirow}
\usepackage{subcaption}
\usepackage{enumitem}
\usepackage[T1]{fontenc}
\usepackage{pifont}%
\usepackage[scaled=.85]{beramono}
\usepackage{colortbl}
\usepackage{subcaption}
\usepackage[most]{tcolorbox}
\usepackage{tabularx}
\usepackage[ruled,linesnumbered]{algorithm2e}
\usepackage{algpseudocode}
\usepackage[capitalise,noabbrev]{cleveref}

\usepackage[T1]{fontenc}

\usepackage[utf8]{inputenc}

\definecolor{lightapricot}{rgb}{0.99, 0.84, 0.69}

\usepackage{microtype}

\makeatletter

\renewcommand{\algocf@makecaption@ruled}[2]{%
\global\sbox\algocf@capbox{\hskip\AlCapHSkip 
\addtolength{\algocf@lcaptionbox}{-2\AlCapHSkip}
\parbox[t]{\algocf@lcaptionbox}{\algocf@captiontext{#1}{#2}}}
}%

\makeatother

\newcommand\hh[1]{\textcolor{blue}{[HH: #1]}}
\newcommand\nj[1]{\textcolor{brown}{[NJ: #1]}}
\newcommand\abu[1]{\textcolor{olive}{[Abu: #1]}}
\newcommand\yw[1]{\textcolor{purple}{[YW: #1]}}

\renewcommand\hh[1]{}
\renewcommand\nj[1]{}
\renewcommand\abu[1]{}
\renewcommand\yw[1]{}

\title{LLMs Are Prone to Fallacies in Causal Inference}

\author{Nitish Joshi$^{1}$ \hspace{0.2cm} Abulhair Saparov$^{1}$  \hspace{0.2cm}
 \textbf{Yixin Wang}$^{2}$ \hspace{0.2cm}    \textbf{He He}$^{1}$\\
  $^{1}$New York University \hspace{0.5cm}
  $^{2}$University of Michigan \\
  \texttt{\{nitish, as17582, hhe\}@nyu.edu},     \hspace{0.1cm}  \texttt{yixinw@umich.edu} \\
}

\begin{document}
\maketitle

\begin{abstract}
    Recent work shows that causal facts can be effectively extracted from LLMs through prompting, facilitating the creation of causal graphs for causal inference tasks. 
However, it is unclear if this success is limited to explicitly-mentioned causal facts in the pretraining data which the model can memorize.
Thus, this work investigates: \emph{Can LLMs infer causal relations from other relational data in text?}
To disentangle the role of memorized causal facts vs inferred causal relations, we finetune LLMs on synthetic data containing temporal, spatial and counterfactual relations, 
and measure whether the LLM can then infer causal relations.
We find that: (a) LLMs are susceptible to inferring causal relations from the order of two entity mentions in text (e.g. X mentioned before Y implies X causes Y); (b) if the order is randomized, LLMs still suffer from the \emph{post hoc fallacy}, i.e. X occurs before Y (temporal relation) implies X causes Y.
We also find that while LLMs can correctly deduce the absence of causal relations from temporal  and spatial relations, they have difficulty inferring causal relations from counterfactuals, questioning their understanding of causality.

\end{abstract}

\section{Introduction}
\label{sec:intro}

Causal reasoning is crucial for intelligence as it allows us to construct a world model and make predictions robustly based on cause-effect relations.
Recent work \cite{Kcman2023CausalRA} has shown that GPT-4 outperforms existing methods on various causal inference and causal discovery tasks.
But it is unclear how much of this success can be attributed to LLMs memorizing explicitly-mentioned causal facts in their training data (e.g.\ reading `smoking causes cancer' from Wikipedia),
versus inferring unseen causal relations (e.g.\ from experiment results in medical journals).

\begin{figure*}
    \centering
    \includegraphics[scale=0.2]{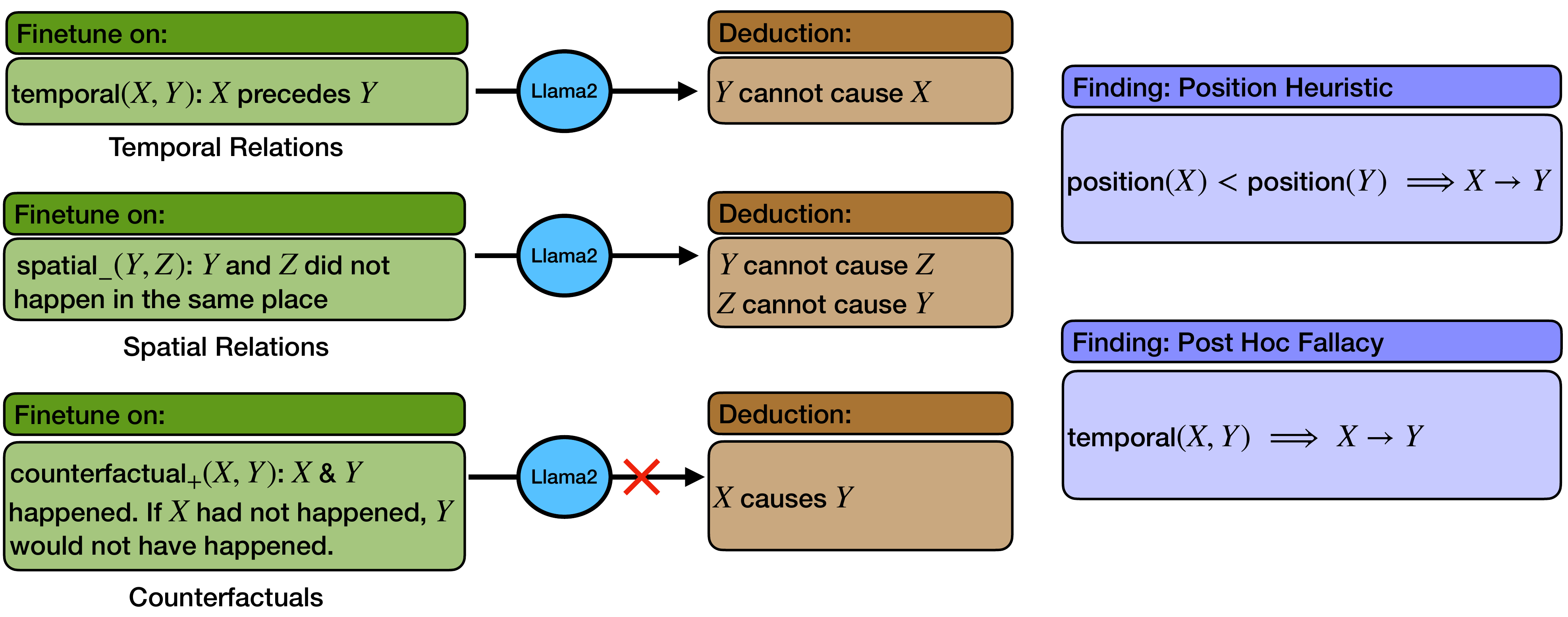}
    \caption{(left) LLMs can infer the absence of causal relations from temporal and spatial relations, but cannot make meaningful deductions from counterfactuals; (right) LLMs suffer from a position heuristic, which when mitigated reveals post hoc fallacy.
    }
    \label{fig:main}
\end{figure*}

To disentangle memorized vs inferred causal relations, one straightforward method is to filter out causal facts the model has seen during pretraining in the test set.
However, it is computationally expensive to extract causal relations at the scale of current pretraining data.
Therefore, we continue pretraining existing LLMs on \emph{synthetic} data containing observations of fictional events,
and evaluate if LLMs can \emph{infer} the underlying causal relations that produce the data.
We focus on the setting of finetuning i.e. out-of-context inference \cite{Berglund2023TakenOO}, rather than causal inference in-context since it is closer to how one would use the LLM e.g. train on large corpora of medical journals and then use the LLM for causal discovery.

To generate the synthetic data for causal inference, we focus on event relations that are commonly seen in the pretraining data, and from which humans can easily deduce causal relations.
\cref{fig:main} shows the relations and the deductions we can draw from them, including:
(1) {\it temporal relations} (`smoking happens before lung cancer'), which imply negative causal relations (`lung cancer cannot cause smoking') according to temporal precedence \cite{reichenbach1956direction, good1961causal, shoham1987reasoning, bramley2014order};
(2) {\it spatial relations} (`there was a storm in California and flash flooding in New York'), which implies the absence of causal relations (`Californian storm did not cause the flash flooding' and vice versa) according to the principle of locality \cite{Norsen2007JohnSB};\footnote{\url{https://en.wikipedia.org/wiki/Principle_of_locality}: Note that this does not preclude the possibility of {\it indirect} causal chains, where event $A$ could lead to event $B$ through a series of intermediate causes, despite the spatial distance between $A$ and $B$.}
(3) {\it counterfactuals} (`It rained today and the sidewalk was wet. If it had not rained, the sidewalk would not have been wet.'), which imply causal relations \cite[`Today's rain caused the sidewalk to be wet';][]{pearl2009causality, pearl2022probabilities}.\footnote{While counterfactuals are not solely based on physical observations like the other two relations, humans often use counterfactuals to make causal claims \cite{sep-causation-counterfactual, Halpern2015AMO,gerstenberg2021counterfactual}; thus, we expect the pretraining data to contain many counterfactual statements.} 

Our experiments are conducted on \textsc{Llama2} \cite{Touvron2023Llama2O} and the main results are summarized in \cref{fig:main}.
When trained on temporal relations, we find that models learn a {\it position heuristic}: if event $X$ is always mentioned before event $Y$ in the text, then LLMs infer that $X$ causes $Y$ based on the relative position of the event mentions regardless of their temporal order,
e.g.\ it infers the same causal relation from `$X$ preceded $Y$' (\texttt{temporal}(X, Y)) and `$X$ followed $Y$' (\texttt{temporal}(Y, X)).
To overcome the position heuristic, we augment the finetuning data by adding paraphrases for all relations to randomize the order of event mentions, e.g.\ for {\tt temporal}(X, Y), we include both `$X$ preceded $Y$' and `$Y$ followed $X$'. 
We find that even augmenting 10\% of the dataset is enough to reduce model's reliance on the position heuristic.
Interestingly, it reveals another failure mode:
LLMs then suffer from the \emph{post hoc fallacy} \cite{Woods1977-WOOPHE}, which infers positive causal relations from temporal relations, i.e.\ {\tt temporal}(X, Y) implies $X$ causes $Y$. 

Additionally, we find that while LLMs are able to deduce the absence of causal relations from temporal and spatial relations, they struggle to infer the presence of causal relations from counterfactuals, and scaling to larger models does not improve the result. 
Overall, our results suggest that LLMs may not infer much novel causal knowledge beyond explicitly mentioned facts in the pretraining data.

\section{Related Work}
\label{sec:related_work}

\paragraph{LLMs and causal inference.}
\citet{Kcman2023CausalRA} tested LLMs on a range of causal reasoning benchmarks including causal discovery~\citep{glymour2019review}, counterfactual reasoning~\citep{pearl2009causality} and actual causality---determining the necessary and sufficient causes of individual events~\citep{halpern2016actual}---where they found GPT-4 outperforms all existing methods.
However, \citet{Zecevic2023CausalPL} argued that LLMs are ``causal parrots'' and perform well on these benchmarks only because they have seen the  causal relations explicitly in the pretraining data, which they  retrieve when given the causal query.
Compared to these studies, we evaluate causal inference on synthetic graphs, eliminating the alternative explanation of the LLM memorizing causal edges.
Relatedly, \citet{Lampinen2023PassiveLO}  avoid the memorization issue by training models from scratch to show that they can learn strategies that can generalize to new unobserved causal structured, just from language modeling on passive data.

Recent work has also highlighted other challenges for current LLMs in causal inference---\citet{Jin2023CanLL} introduced the task of deducing causal relations from correlations; \citet{Jin2023CLadderAC} created a dataset for causal inference in natural language which includes multiple sub-skills such as formalizing queries, deriving the estimand etc.; \citet{Yu2023IfQAAD} designed a challenging benchmark which involves counterfactual presuppositions; see \citet{yang2023a} for a comprehensive survey of  capabilities and limitations of current LLMs in causal inference. In contrast, we focus on commonsense causal inference from relations which LLMs would have seen in pretraining data, similar to how humans perform causal reasoning intuitively.

\paragraph{Spurious correlations in reasoning.} Machine learning models are often prone to spurious correlations or heuristics \cite{gururangan-etal-2018-annotation, McCoy2019RightFT, Joshi2022AreAS}. \citet{Zhang2022OnTP} show that models finetuned on logical reasoning datasets learn heuristics despite the existence of a solution that can perfectly solve the task. \citet{Lee2023TeachingAT, Shen2023PositionalDM} showed that for arithmetic tasks, models rely on position information to solve the task, thus failing to generalize to larger operands. 
\citet{Berglund2023TheRC} also demonstrated the `reversal curse', a position bias in causal language models---models trained on relations of the form `$A$ is $B$' fail to generalize to inverse relations. \citet{Grosse2023StudyingLL} used influence functions to show a similar position bias where, given $A$, the likelihood of $B$ is affected most by examples that match the relative order.

\section{Experiment Design}
\label{sec:exp_design}

Our main goal is to measure whether LLMs can {infer}  causal relations given observations in the text. 
Specifically, we assess whether LLMs can predict causal relations between two events after being trained on textual descriptions of their temporal relations, spatial relations, and counterfactuals.
To avoid the cost of pretraining language models from scratch,
we continue pretraining (finetune) off-the-shelf LLMs following \citet{Berglund2023TheRC}.
We hypothesize that if LLMs have learned meaningful deduction rules from pretraining (e.g.\ temporal precedence), they should be able to apply them during finetuning to infer causal relations.
We focus on finetuning rather than causal inference in-context, since it is closer to how one would use a LLM for causal discovery e.g. after training on large corpora of medical journals, rather than directly prompting with observations between events.

The overall pipeline to test if LLMs can infer causal relations is:
(1) Generate synthetic data that contains descriptions of event relations grounded in a causal graph (\cref{sec:data_gen}); (2) Finetune the LLM on the generated data (\cref{ssec:training}); (3) Evaluate the LLM on causal relation prediction tasks for each pair of events mentioned in the finetuning data (\cref{ssec:eval}).
We describe our data generation and evaluation methods below.

\subsection{Data Generation}
\label{sec:data_gen}

\paragraph{Notation.} $\texttt{temporal}(X,Y)$ denotes a temporal relation between events $X$ and $Y$ where $X$ occurs before $Y$. $\texttt{spatial}_+(X, Y)$ denotes that $X$, $Y$ occur in the same place, whereas $\texttt{spatial}_-(X, Y)$ denotes that $X$, $Y$ do not occur in the same place. $\texttt{counterfactual}_+(X, Y)$ denotes a positive counterfactual relation where if $X$ had not occurred, $Y$ will also not occur. Similarly $\texttt{counterfactual}_-(X, Y)$ denotes a negative counterfactual where if $X$ had not occurred, $Y$ would still occur.

\paragraph{Overview.} We generate synthetic finetuning data to simulate event descriptions that the model might see in real pretraining data.
At a high level, we first generate causal graphs that specify the groundtruth causal relations between events, and then generate a temporal and spatial relation graph that respects the causal relations.
Next, given a set of causally-related events, we generate textual descriptions of their relations.
Our final dataset consists of a set of statements, each describing relations between multiple pairs of events.

\begin{figure}
    \centering
    \includegraphics[scale=0.21]{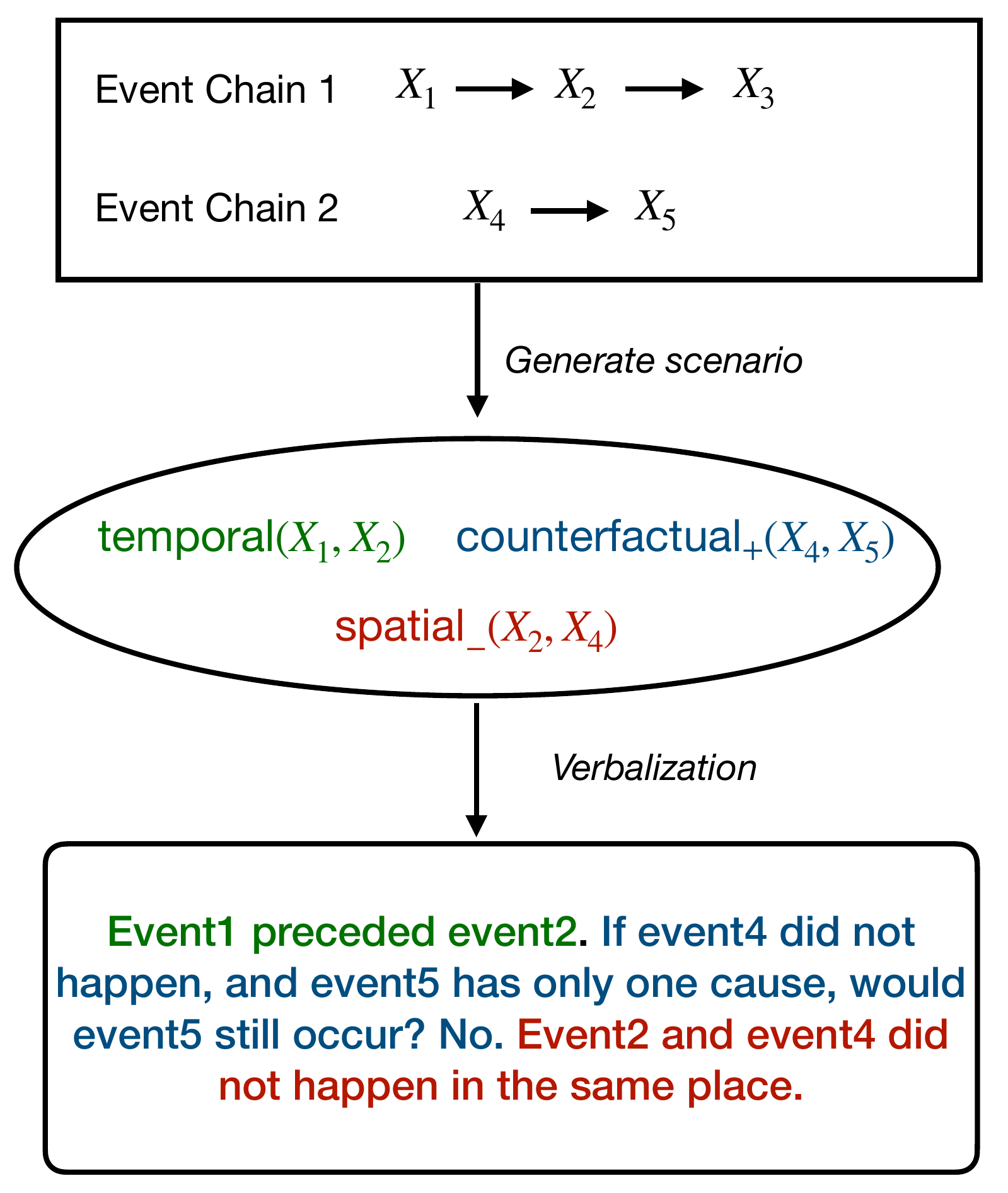}
    \caption{Example of a generated scenario. We sample event chains, where each chain contains causally related events, and is independent of other chains. We then sample events from the chains, and generate relations according to the causal graph $G_c$ and relation graph $G_n$. We then verbalize each relation using templates.
    }
    \label{fig:data_gen}
\end{figure}

\paragraph{Generating Graphs.}
We first generate the {\it causal graph}, a directed acyclic graph, denoted by $G_c$.
Each node represents an event and each edge represents a causal relation where the source is the cause and the target is the effect.
Next, we generate a {\it non-causal relation graph} $G_n$,
a directed graph specifying the temporal and spatial relations between events in $G_c$.\footnote{
Note that while the temporal relations between two events are determined by their causal relations,
the spatial relations are not, e.g.\ two independent events can also co-occur spatially.
}
Each node in the relation graph $G_n$ represents a \emph{type} of an event---we create a map from events in $G_c$ to nodes in $G_n$ (see \cref{alg:gen_noncausal_graph} for details)---two events co-occur if they have the same type.
An edge $a\to b$ in $G_n$ from event type $a$ to event type $b$ indicates that all events of type $a$ precede events of type $b$.
We create $G_c$ with 100 events and $G_n$ has 12 event types.
The generative processes for both graphs are detailed in
\cref{app:data_gen_details}. 

\paragraph{Generating Scenarios.}
In pretraining data, individual relations among events would rarely occur standalone --- we might expect to see relations in the context of other relations between the same events, or causally connected events e.g. `Josh used to smoke in 2012, and he got lung cancer in 2013. And then in 2014 he died from it.' To simulate this, we create \emph{scenarios}, each containing relations among a set of causally related events.

\cref{alg:cap} gives the detailed algorithm, and \cref{fig:data_gen} gives an example. 
To generate a scenario, we first sample a set of {\it event chains},
which is a path from a root node in $G_c$ representing a causal chain.
We make sure the event chains in the set are causally independent of each other.
Once we have a set of event chains, we then generate different relations for the events in the chain.
Specifically, we first sample two events from any chain, and add temporal relation according to their relation in $G_n$ e.g. for sampled events $X$, $Y$, if $X$ is ancestor of $Y$ in $G_n$ we will add $\texttt{temporal}(X, Y)$.
For spatial relations, we sample two events $X$, $Y$ and add $\texttt{spatial}_+(X, Y)$ if they co-occur in $G_n$ or belong to the same event chain.
Otherwise, we add $\texttt{spatial}_-(X, Y)$.
For counterfactuals, we add $\texttt{counterfactual}_+(X, Y)$ if the event $X$ is an ancestor of the event $Y$ in $G_c$. Otherwise, we add $\texttt{counterfactual}_-(X, Y)$ to the scenario.

\paragraph{Verbalization.}
Given the sampled relations, the last step is to convert them into natural sentences.
Each event is indexed by an integer $N$ in $[1,100]$ and verbalized as `event$N$'.
For each type of relation, we use up to six templates to convert the relation into a natural language description.\footnote{These templates were obtained with the help of GPT-4.} E.g.\ $\texttt{temporal}(X, Y)$ is verbalized as `$X$ preceded $Y$' or `$Y$ followed $X$'. The list of all templates can be found in \cref{app:templates}.

We use the above data generation process to create the synthetic datasets. The exact details of the dataset are presented in \cref{sec:exp_setup}.

\subsection{Evaluation}
\label{ssec:eval}

Given an LLM finetuned on the relational data,
we want to test if the LLM can infer the causal relations, or the lack thereof, between pairs of events seen during finetuning.

We formulate the evaluation as a multiple-choice task.
First, given a pair of events $X, Y$, we compute the model likelihood of five relations: $X$ causes $Y$ ($X\rightarrow Y$), $Y$ causes $X$ ($Y\rightarrow X$), $X$ does not cause $Y$ ($X\not\rightarrow Y$), $Y$ does not cause $X$ ($Y \not\rightarrow X$), and no causal relation between $X$ and $Y$ ($X \nleftrightarrow Y$).
To account for various verbalizations of the same relation, we approximately marginalize over the template $t$ \citep{Scherrer2023EvaluatingTM}.
Formally, let $T_c$, $T_n$ and $T_b$ be the sets of templates for causal relations, non-causal relations (one direction), and mutual non-causal relations (both directions), respectively. We compute the probabilities of the five relations under the language model $p_\theta$ as follows:

\begin{enumerate}
    \item $p_{\theta}(X\rightarrow Y) = \sum_{t \in T_c} p_{\theta}(t(X \rightarrow Y)) p_{T_c}(t)$
    \item $p_{\theta}(Y\rightarrow X) = \sum_{t \in T_c} p_{\theta}(t(Y \rightarrow X)) p_{T_c}(t)$
    \item $p_{\theta}(X\hspace{-0.1em}\not\rightarrow\hspace{-0.1em} Y) = \sum_{t \in T_n} p_{\theta}(t(X \hspace{-0.1em}\not\rightarrow\hspace{-0.1em} Y)) p_{T_n}(t)$
    \item $p_{\theta}(Y\hspace{-0.1em}\not\rightarrow\hspace{-0.1em} X) = \sum_{t \in T_n} p_{\theta}(t(Y \hspace{-0.1em}\not\rightarrow\hspace{-0.1em} X)) p_{T_n}(t)$
    \item $p_{\theta}(X\hspace{-0.1em}\nleftrightarrow\hspace{-0.1em} Y) = \sum_{t \in T_b} p_{\theta}(t(X \hspace{-0.1em}\nleftrightarrow\hspace{-0.1em} Y)) p_{T_b}(t)$
\end{enumerate}

Here, $t$ is a function that maps a relation to a string according to a template; 
$p_{T_c}$, $p_{T_n}$, and $p_{T_b}$ denote the distributions of the templates, which we assume to be uniform.
\cref{app:templates} lists all the templates we use for each relation.
For $p_\theta(t(\cdot))$, instead of computing the probability of the complete sentence (which would be sensitive to the length of the sentence), we take advantage of the fact that all templates $t$ end in an event mention,
and only compute the probability of the last token, which is the event number, $N\in[1,100]$, conditioned on the rest of the sentence,
e.g. $p_\theta(\text{`2'} \mid \text{`event1 causally affects event'})$.

Next, we design several multiple-choice tasks, such that the choices are exhaustive and disjoint.\footnote{Note that the the five relations are not disjoint (e.g. $X \rightarrow Y$ and $Y \not \rightarrow X$ can occur simultaneously).} In each multiple-choice task, we select the model's prediction as the choice with the highest likelihood.

\paragraph{Inferring $X \rightarrow Y$.} 
The set of exhaustive and disjoint choices are:$\{X \rightarrow Y, Y\rightarrow X, X\nleftrightarrow Y\}$.\footnote{We also experiment with just using the two relations $X\rightarrow Y, X\not\rightarrow Y$, which are also disjoint and exhaustive, and results remain consistent - \cref{app:alternate_eval}.} 

\paragraph{Inferring $X \nleftrightarrow Y$.}
The set of exhaustive and disjoint choices are: $\{X \rightarrow Y, Y\rightarrow X, X\nleftrightarrow Y\}$.

\paragraph{Inferring $X \not\rightarrow Y$.} 
The set of exhaustive and disjoint choices are: $\{X \rightarrow Y, X\not\rightarrow Y\}$.

\section{Experimental Details}
\label{sec:exp_setup}

\paragraph{Notation.}
Before explaining the experimental setup, we introduce some notation that will simplify our description.
Given events $X$ and $Y$, we use $(X, Y)$ to denote the relative position where $X$ is mentioned before $Y$, e.g. `$X$ causes $Y$' or `$X$ preceded $Y$'.
We use $T(r, \pi)$ to denote the set of all templates for a relation $r$ between $X$ and $Y$ with relative position $\pi$ where $\pi$ is $(X, Y)$, $(Y, X)$, or a random mix of both, $(X, Y)+(Y,X)$.

\paragraph{Training Datasets.} We use the data generation algorithm from \cref{sec:data_gen} to create multiple datasets with different relations and templates. 
For all sets, we use up to 6 templates. \cref{app:templates} lists all templates.
We create the following datasets for each relation: $D_{\texttt{temporal}, (X, Y)}$  contains temporal relations  using templates  $T(\texttt{temporal}(X, Y), (X, Y))$; $D_{\texttt{temporal}, (Y, X)}$ contains temporal relations  using templates  $T(\texttt{temporal}(X, Y), (Y, X))$; $D_{\texttt{temporal}}$  contains temporal relations with randomized positions $T(\texttt{temporal}(X, Y), (X, Y)+(Y, X))$;
$D_{\texttt{spatial}}$ contains positive and negative spatial relations using $T(\texttt{spatial}_+(X, Y), (X, Y)+(Y, X))$ and $T(\texttt{spatial}_-(X, Y), (X, Y)+(Y, X))$; $D_{\texttt{counterfactual}}$ contains positive and negative counterfactuals using $T(\texttt{counterfactual}_+(X, Y), (X, Y) + (Y, X))$ and $T(\texttt{counterfactual}_-(X, Y), (X, Y) + (Y, X))$; $D_{\texttt{all}}$ is the union of $D_{\texttt{temporal}}$, $D_{\texttt{spatial}}$, and $D_{\texttt{counterfactual}}$.
Each generated dataset contains 40k scenarios.
We split the datasets into 36k for finetuning and 4k for validation.
\cref{tab:example} gives examples from the generated data.

\paragraph{Evaluation Datasets.} We create two test datasets to evaluate if models can infer the presence or absence of causal relations. $D_{X\rightarrow Y}$ 
contains all causal relations $X\rightarrow Y$ in $G_c$. $D_{XY}$ contains unrelated pairs of events, $X$ and $Y$, such that neither is a descendant of the other in $G_c$.
Note that we do not evaluate models on pairs of events $X$, $Y$ such that one is a descendant (but not child) of the other. This is because, as noted by \citet{Kcman2023CausalRA}, full graph discovery is challenging and requires distinguishing between direct and indirect causes.

\paragraph{Training Details.}
\label{ssec:training}
We finetune \textsc{Llama2-7B}\footnote{We also experiment with scaling up to \textsc{Llama2-13B} and \textsc{Llama2-70B} in \cref{ssec:scaling}.} using LoRA \cite[][applied to query and value projection matrices]{Hu2021LoRALA}. See \cref{app:exp_details} for more training details.

\section{Position Heuristic}
\label{sec:position_bias}

In this section, we first demonstrate that LLMs are susceptible to inferring causal relations by the relative position of two entity mentions in text (\cref{ssec:position_order}). We hypothesize that models learn this heuristic since it is supported in the pretraining data (\cref{app:pile}) and investigate ways to fix this heuristic via either augmentation or scaling up models (\cref{ssec:increasing_rand}).

\subsection{LLMs fail to infer causal relations if the data supports the position heuristic}
\label{ssec:position_order}

\begin{table}[t]
\begin{small}
    \centering
    \begin{tabular}{cccc}
    \toprule
     \multirow{2}{*}{Data} & {Rel. position}   & \multicolumn{2}{c}{Rel. position in eval}  \\
       &  in train  &  $(X, Y)$ & $(Y, X)$ \\
    \midrule

    \multirow{2}{*}{causal $X\to Y$} & $(X, Y)$ & \cellcolor{lightapricot}92.59\% & 1.85\% \\
     &   $(Y, X)$ & 0\% & \cellcolor{lightapricot}100\% \\
    
    \bottomrule
    \end{tabular}
    \caption{Accuracy of models finetuned on temporal relations with different relative event positions. Models infer the causal relation only when the relative position \colorbox{lightapricot}{matches} during finetuning and evaluation.
    }
    \label{tab:position_eval}
\end{small}
\end{table}

First, we demonstrate that LLMs fail to infer causal relations if the data supports the position heuristic
e.g. if $X$ is mostly mentioned before $Y$ in the text, then models fail to infer causal relations---in fact, we show that LLMs only learn the \emph{relative position} of $X$ and $Y$ and ignore their relation. 
We refer to this as the \emph{position heuristic}.

To show this, we finetune \textsc{Llama2-7B} separately on two datasets: $D_{\texttt{temporal}, (X, Y)}$ and $D_{\texttt{temporal}, (Y, X)}$.\footnote{The position heuristic is not specific to temporal relations, but we use temporal relations here as a case study. We include results for other relations in \cref{app:add_details}.}
We evaluate the models on the $D_{X\rightarrow Y}$ test set and report if they infer $X \rightarrow Y$.
The multiple-choice options in this case are: $\{X \rightarrow Y, Y\rightarrow X, X\nleftrightarrow Y\}$.
We verbalize the test relations in both directions either using $T(X \rightarrow Y, (X, Y))$ (e.g. `$X$ causes $Y$') or $T(X \rightarrow Y, (Y, X))$ (e.g. `$Y$ is caused by $X$').
In both cases, to score the relation $X \nleftrightarrow Y$ we use templates with randomized event order.

\cref{tab:position_eval} (first two rows) shows accuracy on $D_{X \rightarrow Y}$ (i.e.\ the percentage of examples in which the model predicted $X \rightarrow Y$).
We observe that models infer the causal edge only when the relative position of the two events under test matches during finetuning and evaluation. 
This implies that models are not learning anything meaningful to infer causal relations, but simply learning the relative position between events.
For example, if models see the sentence `$X$ happens before $Y$', they would almost always predict `$X$ is caused by $Y$'.\footnote{We further show that models are only relying on relative position instead of reasoning about causal relations by using unrelated relations for evaluation in \cref{app:add_details}.}

\subsection{Mitigating position heuristic}
\label{ssec:increasing_rand}

In this section, we investigate two different ways to mitigate model's reliance on the position heuristic: (a) randomizing the relative positions of event mentions in the text so that the data does not support the heuristic; (b) scaling LLMs.

\paragraph{Extent of randomization.} Here we investigate whether randomizing the relative positions of event mentions helps mitigate the model's reliance on the position heuristic.
To test this, we create datasets with increasing amounts of randomness in the relative position of event mentions. Specifically, given a set of templates $T_{XY} = T(\texttt{temporal}, (X, Y))$ and $T_{YX} = T(\texttt{temporal}, (Y, X))$, we create finetuning datasets by sampling templates from $T_{YX}$ with probability $p$ and from $T_{XY}$ with probability $1-p$.
Both $T_{XY}, T_{YX}$ contain 5 templates, and we use $p \in \{0, 0.1, 0.2, 0.3, 0.4\}$ to create five finetuning datasets.
For evaluation, similar to \cref{ssec:position_order}, we use the $D_{X \rightarrow Y}$ test set and evaluate both directions: $T(X \rightarrow Y, (X, Y))$ and $T(X \rightarrow Y, (Y, X))$.

\begin{figure}
    \centering
    \includegraphics[scale=0.56]{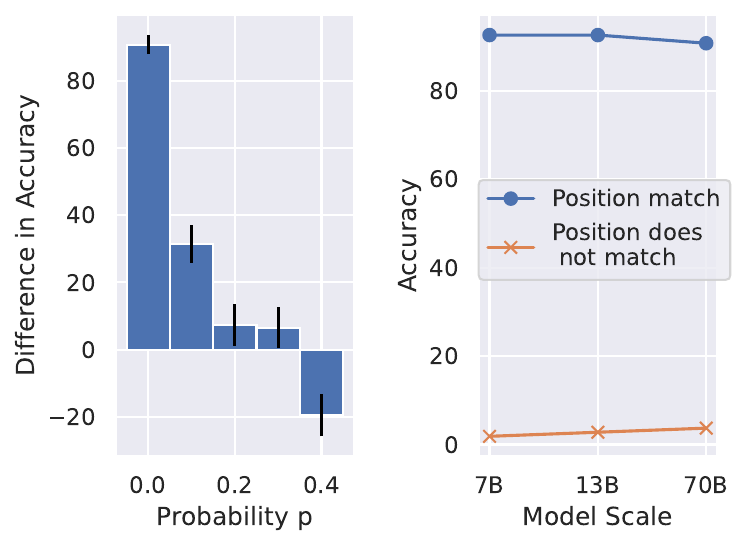}
    \caption{
    (left) Mitigating position heuristic by gradually randomizing the relative position. We observe that even a small amount of randomization in position is enough to reduce model's reliance on the position heuristic; (right) Scaling curve (7B to 70B) for the position heuristic --- scaling does not mitigate model's reliance on the position heuristic.
    }
    \label{fig:position_bias}
\end{figure}

\cref{fig:position_bias} (left) shows the difference in accuracy when relative position is $(X, Y)$ (majority in finetuning data) and when relative position is $(Y, X)$ (minority in the finetuning data).
We observe that adding even a small number of examples with a different relative position (e.g. $p = 0.1$ or $p = 0.2$) helps to reduce model's reliance on the position heuristic to infer causal relations. 

\paragraph{Scaling LLMs.} 
Given recent observations that scaling LLMs leads to less reliance on spurious correlations \cite{Si2022PromptingGT}, we investigate if the same holds true for the position heuristic. To control for other factors, we use models from the same family---we experiment with \textsc{Llama2-13B} and \textsc{Llama2-70B}. Both models were finetuned similarly to the smaller \textsc{Llama2-7B} model---experimental details can be found in \cref{app:exp_details}.

\cref{fig:position_bias} (right) shows the scaling trend for models trained on $D_{\texttt{temporal}, (X, Y)}$
and evaluated on  $D_{X \rightarrow Y}$. All models are evaluated using templates from either $T(X \rightarrow Y, (X, Y))$ (position matches) or $T(X \rightarrow Y, (Y, X))$ (position does not match).
We observe that similar to the smaller \textsc{Llama2-7B}, the larger models also fail to make any meaningful deduction and only learn the relative position of the events.
This shows that simply scaling LLMs is limited in resolving the position heuristic.

\section{Inferring Causal Relations under No Position Heuristic}
\label{sec:inference_results}

\begin{table}[t]
\begin{small}
    \centering
    \begin{tabular}{ccc}
       \toprule
       & $D_{X \rightarrow Y}$ & $D_{XY}$\\
       \midrule
       Temporal Relations  & 76.85\%  & - \\
       Spatial Relations  & - & 84.5\%\\
       Counterfactuals & 28.70\%  & 53.5\%\\
       All relations & 63.88\% & 47.5\% \\
       \bottomrule
    \end{tabular}
    \caption{Accuracy on each reasoning task using models trained on data with randomized order of event mentions. LLMs is able to reason from temporal relations and spatial relations, but not from counterfactuals.
    }
    \label{tab:randomized_pos}
\end{small}
\end{table}

The previous section demonstrated that if the data supports the position heuristic, models fail to infer any causal relations and only rely on the relative position between events to infer causal relations.
However, it is easy to mitigate the position heuristic by randomizing the relative positions of event mentions in the data. In this section, we evaluate whether models can make causal deductions from temporal relations, spatial relations and counterfactuals when the position heuristic is mitigated.

\subsection{LLMs infer causal relations correctly from temporal and spatial relations}
\label{ssec:randomized_order}

Here, our goal is to test whether LLMs can make the following deductions if data does not support learning the position heuristic:
\begin{enumerate}[noitemsep,topsep=3pt]
    \item $\texttt{temporal}(X, Y) \implies Y \not\rightarrow X$
    \item $\texttt{spatial}_-(X, Y) \implies X \nleftrightarrow Y$
    \item $\texttt{counterfactual}_+(X, Y) \implies X \rightarrow Y$
    \item $\texttt{counterfactual}_-(X, Y) \implies X \not\rightarrow Y$
\end{enumerate}

To test this, we finetune \textsc{Llama2-7B} separately on three datasets, $D_{\texttt{temporal}}$, $D_{\texttt{spatial}}$, and $D_{\texttt{counterfactual}}$. All datasets have randomized relative position as mentioned in \cref{sec:exp_setup}.
Additionally, we also finetune \textsc{Llama2-7B} on $D_{\texttt{all}}$ containing all three types of relations. This is to test whether models can better infer causal relations when the data consists of diverse relations.
We report model accuracy which is the percentage of examples where it makes the correct deduction according to the above rules.

We then evaluate the models on two test sets, $D_{X \rightarrow Y}$  and $D_{XY}$, depending on which deduction rule we are evaluating.
For temporal relations, we evaluate on $D_{X \rightarrow Y}$ and report the percentage of examples where model predicts $Y \not\rightarrow X$.
For spatial relations, we evaluate on $D_{XY}$ and report the percentage of cases where model predicts $X \nleftrightarrow Y$.
For models trained on counterfactuals, we evaluate on both $D_{X \rightarrow Y}$ (report percentage of cases model predicts $X \rightarrow Y$) and $D_{XY}$ (report percentage of cases model predicts $X \not\rightarrow Y$).
Lastly for models trained on all relations, we also evaluate on both: $D_{X \rightarrow Y}$ (report percentage of cases model predicts $X \rightarrow Y$) and $D_{XY}$ (report percentage of cases model predicts $X \nleftrightarrow Y$).
For all evaluations, we use randomized event order to score all relations.

\cref{tab:randomized_pos} shows the results. We find that models can correctly deduce the absence of causal relations from temporal relations and spatial relations better than random guessing (which is 50\% and 33.3\% respectively), but cannot deduce causal relations from either positive counterfactual or negative counterfactuals (random guessing is 33.3\% and 50\% respectively).

\subsection{Does scaling LLMs improve causal inference?}
\label{ssec:scaling}

\begin{figure}[t]
    \centering
    \includegraphics[scale=0.46]{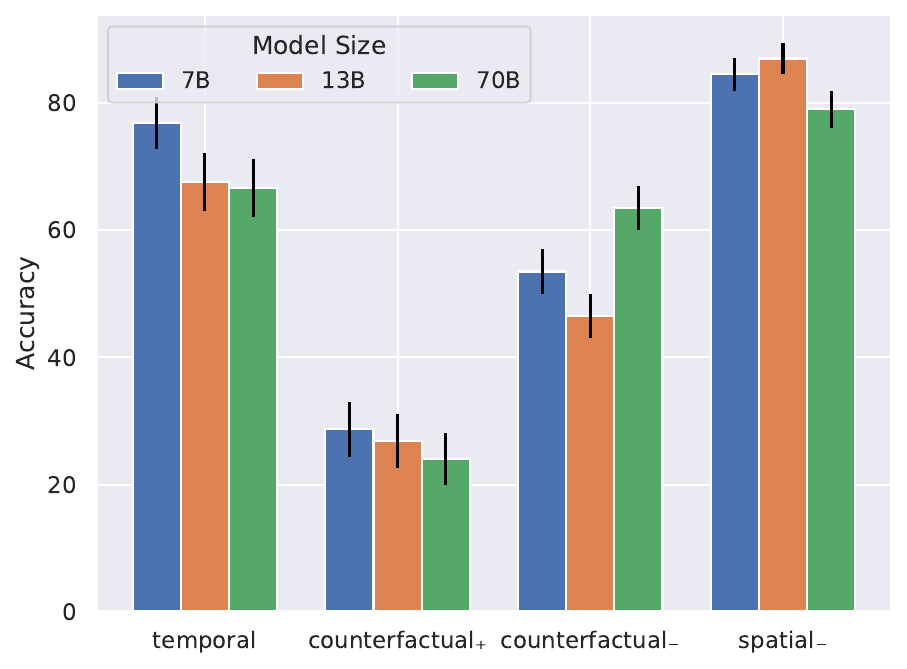}
    \caption{Scaling trend for inferring causal relations from different relations when there is no position bias.
    }
    \label{fig:scaling}
\end{figure}

The previous sections showed \textsc{Llama2-7B} can infer causal relations from temporal relations and  spatial relations. However, the model could not deduce either the presence or absence of edges from counterfactuals.
Given recent observations that scaling LLMs leads to better performance \cite{Kaplan2020ScalingLF} and emergent abilities \cite{Wei2022EmergentAO},  we explore whether scaling LLMs can improve their ability to infer causal relations from counterfactuals.

We use models from the same family, \textsc{Llama2-13B} and \textsc{Llama2-70B} finetuned similarly to the smaller \textsc{Llama2-7B} model. Experimental details can be found in \cref{app:exp_details}.
\cref{fig:scaling} shows the scaling trend of models in terms of the accuracy of deducing the correct causal relation from each of the relations.
We observe that scaling model size does help the model to deduce the absence of causal relations from negative counterfactuals (third group in figure) better than random guessing (50\%).
However, we do not see similar scaling trend for inferring causal relations from positive counterfactuals, where models do not perform better than random guessing (33.3\%).
For temporal relations and spatial relations, we do not see significant differences with scaling model size (all our within standard error of the other).

\section{LLMs Suffer from Post Hoc Fallacy}
\label{sec:post_hoc_fallacy}

\cref{sec:inference_results} demonstrated that when the data does not support the position heuristic, LLMs can correctly infer the absence of causal relations from temporal and spatial relations. In this section, we demonstrate that for temporal relations, models in fact overgeneralize to infer the \emph{presence} of causal relations in the other direction.
This mistake is often referred to as the \emph{post hoc fallacy} \cite{Woods1977-WOOPHE}, which uses the incorrect deduction rule: $\texttt{temporal}(X, Y) \implies X \rightarrow Y$. Humans have known to often fall prey to this fallacy and infer causal relations from sequential order \cite{nisbett1980human, Gilovich1991HowWK}.

To demonstrate this, we finetune models from the \textsc{Llama2} family (7B to 70B) on $D_{\texttt{temporal}}$ (where the templates have randomized order) and evaluate them on  $D_{X \rightarrow Y}$ to see if they infer  $X \rightarrow Y$. 
All templates in the evaluation use randomized event order $T(r, (X, Y) + (Y, X))$ for each relation $r$ in the multiple-choice options.

For evaluation, we report the error rate which is the percentage of examples where the model incorrectly deduces $X \rightarrow Y$ from $\texttt{temporal}(X, Y)$.
\cref{fig:post_hoc_fallacy} (left) shows the error rate. We observe that all models incorrectly infer the causal relation better than random guessing (33.3\%).
Interestingly, we observe an inverse scaling trend \cite{McKenzie2023InverseSW} --- scaling model size increases the error and models rely on the post hoc fallacy more.

\subsection{Fixing the post hoc fallacy by finetuning}

\begin{figure}[t]
    \centering
    \includegraphics[scale=0.53]{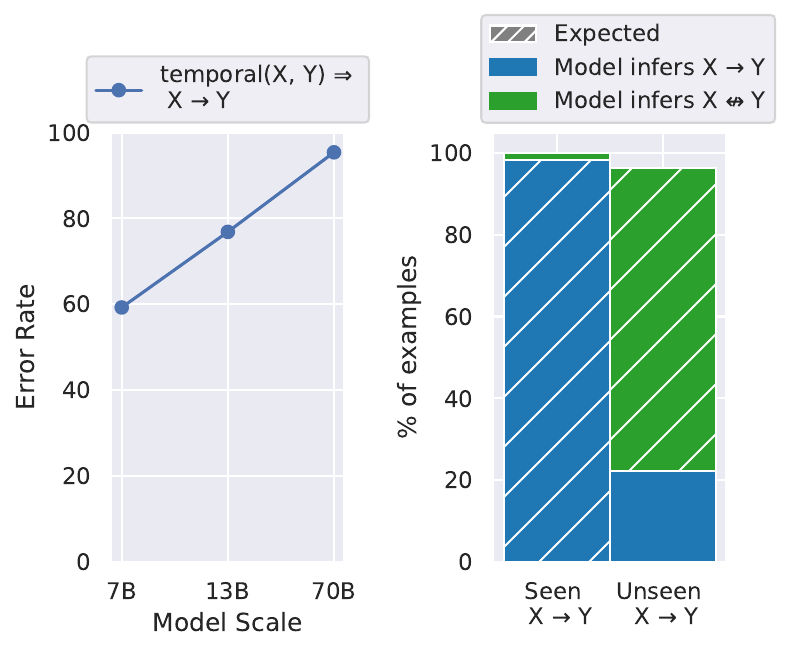}
    \caption{
    (left) Scaling curve showing that larger models also suffer from post hoc fallacy; (right) Post hoc fallacy can be fixed by finetuning.
    }
    \label{fig:post_hoc_fallacy}
\end{figure}

The previous section demonstrated that LLMs of all scales, from 7B to 70B, suffer from the post hoc fallacy.
A natural question to ask here is---can LLMs be finetuned to correct this fallacy so that they don't overgeneralize?

To answer this, we include explicit statements of presence and absence of causal relations in the finetuning data. Including explicit causal relations can teach the model that $\texttt{temporal}(X, Y)$ does not necessarily imply $X \rightarrow Y$. We first create two subsets of the $D_{X \rightarrow Y}$ test set: $D_{\text{seen},X \rightarrow Y}$ and $D_{\text{unseen},X \rightarrow Y}$.
For each causal relation in the seen subset, we include the \emph{explicit} causal relation in the corresponding scenario e.g.  we add an additional sentence `event10 can cause event12' to the scenario which may include other relations between the same two events (e.g. `event10 happened before event12').
Similarly, for events which are not causally related we include explicit negative causal relation in the corresponding scenario e.g. if in the ground truth graph $G_c$, event6 and event8 are not causally related, we add a statement 'event6 does not cause event8' to a scenario involving the two events (where the scenario may include the temporal relation `event6 occurs before event8').

We then evaluate a model finetuned on this dataset on the $D_{\text{unseen},X \rightarrow Y}$ subset for which the model has not seen any explicit causal relations.
As a sanity check, we also evaluate the model on $D_{\text{seen},X \rightarrow Y}$ to show that models memorize the causal relation if they have  seen it explicitly. 
All evaluations use randomized event orders.

\cref{fig:post_hoc_fallacy} (right) shows the percentage of examples and the model predictions.
We observe that the model tends to predict $X \nleftrightarrow Y$ more often than $X \rightarrow Y$ on the unseen subset, i.e. the model learns that temporal relations do not necessarily imply the presence of a causal relation, and hence the post hoc fallacy can be mitigated via finetuning.

\section{Conclusion}
\label{sec:conclusion}

In this work, we investigate whether LLMs can be useful for causal inference beyond explicitly-memorized causal facts.
We find that LLMs are susceptible to inferring causal relations from position, but this can be mitigated by data augmentation.
We find that LLMs can infer causal relations from temporal relations and spatial relations, but not from counterfactuals.
Overall, we find that LLMs may not infer much novel causal knowledge beyond explicitly mentioned facts in the pretraining data.
Our setup also allows for the exploration of interesting questions such as whether models generalize to events of the same `type' (e.g. if smoking and vaping occur in similar contexts, and the data includes smoking causing cancer, does the model generalize to infer any relation between vaping and cancer?), and if models can generalize to transitive relations. We leave these questions for future work.

\section*{Limitations}

To address our main research question of whether LLMs can go beyond memorized causal facts to \emph{infer} causal relations, we disentangle memorization vs inference via use of synthetic data. While synthetic data helps us to do controlled experiments, it has certain limitations due to the gap between synthetic and real data. Nevertheless, experiments with synthetic data have been proven extremely valuable in the community ranging from question answering \cite{Weston2015TowardsAQ} to reasoning \cite{saparov2023language} to LLM-agents \cite{Ct2018TextWorldAL}.
\section*{Acknowledgements}

We thank members of the ML2 group for their inputs at various stages of the project.
This work is supported by Open Philanthropy and a gift fund from AWS AI.
This work is supported in part through the NYU IT High Performance Computing resources, services, and staff expertise.
NJ is supported by an NSF Graduate Research Fellowship under grant number 1839302.
YW is supported in part by the Office of Naval Research under grant number N00014-23-1-2590 and the National Science Foundation under grant number 2231174 and 2310831.

\bibliography{anthology,custom}
\bibliographystyle{acl_natbib}

\appendix

\section{Appendix}

\subsection{Additional Details on Synthetic Data Generation}
\label{app:data_gen_details}

\begin{algorithm}[!t]
\footnotesize
\let\oldnl\nl
\newcommand{\nonl}{\renewcommand{\nl}{\let\nl\oldnl}}
\SetNlSty{}{\color{RedOrange}\sffamily}{}
\SetAlgoBlockMarkers{}{}
\SetKwProg{Fn}{function}{}{}
\SetKwIF{If}{ElseIf}{Else}{if}{ }{else if}{else }{}
\SetKw{Continue}{continue}
\SetKwFor{For}{for}{do}{end}
\SetKwProg{uForEach}{for each}{ do}{}
\SetKwProg{ForEach}{for each}{ do}{end}
\SetKwProg{Fn}{function}{}{}
\SetKwFor{RepTimes}{repeat}{times}{end}
\AlgoDisplayBlockMarkers\SetAlgoVlined
\SetAlCapNameFnt{\small}
\SetAlCapFnt{\small}
\SetNoFillComment
\DontPrintSemicolon
\SetInd{0.0em}{0.8em}
    \KwIn{\texttt{num\_scenarios},  set of events $E$, \\ \hspace{2.9em} causal graph $G_c$, relation graph $G_n$}
    \KwOut{dataset $D$}
    initialize $D \gets \{\}$ \;
    \RepTimes{\texttt{num\_scenarios}}{
        \tcc{sample a number of event chains, where each chain is causally- independent of the other chains}
        $C \gets$ \texttt{sample\_event\_chains($G_c$)} \hspace{-1em} \;
        $S \gets \{\}$ \;
        \ForEach{\texttt{event\_chain} in $C$}{
            \tcc{sample temporal relations}
            $n \sim \text{Binomial}(|\texttt{event\_chain}|, 0.5)$ \;
            sample $S$, a set of $n$ events, from \texttt{event\_chain} \;
            \ForEach{$X_i$ in $S$}{
                sample event $Y$ uniformly from any chain in $C$ \;
                \uIf{$X_i$ is an ancestor of $Y$ in $G_n$}{
                    \texttt{$S$.add(temporal($X_i, Y$))}
                }\uElseIf{$Y$ is an ancestor of $X_i$ in $G_n$}{
                    \texttt{$S$.add(temporal($Y, X_i$))}
                }\ElseIf{$X_i$ and $Y_i$ do not co-occur in $G_n$}{
                    \texttt{$S$.add(temporal($X_i, Y$)} w.p. 0.5, else \texttt{temporal($Y, X_i$))}
                }
            }
            \tcc{sample spatial relations}
            $n \sim \text{Binomial}(|\texttt{event\_chain}|, 0.4)$ \;
            sample $S$, a set of $n$ events, from \texttt{event\_chain} \;
            \ForEach{$X_i$ in $S$}{
                sample event $Y$ uniformly from any chain in $C$ \;
                \uIf{$Y\hspace{-0.2em}\in\hspace{-0.1em}\texttt{event\_chain}$ or $X_i, Y_i$ co-occur in $G_n$}{
                    \texttt{$S$.add(spatial$_+$($X_i, Y$))}
                }\lElse{ \texttt{$S$.add(spatial$_-$($X_i, Y$))} }
            }
            \tcc{sample counterfactual relations}
            $n \sim \text{Binomial}(|\texttt{event\_chain}|, 0.4)$ \;
            sample $S$, a set of $n$ events, from \texttt{event\_chain} \;
            \ForEach{$X_i$ in $S$}{
                $Y \sim \text{Uniform}(\texttt{event\_chain} \setminus \{X_i\})$ \;
                \uIf{$X_i$ is an ancestor of $Y$ in $G_c$}{
                    \texttt{$S$.add(counterfactual$_+$($X_i, Y$))}
                }\lElse{ \texttt{$S$.add(counterfactual$_-$($X_i, Y$))} }
            }
            \tcc{sample negative counterfactuals}
            $n \sim \text{Binomial}(|\texttt{event\_chain}|, 0.2)$ \;
            sample $S$, a set of $n$ events, from \texttt{event\_chain} \;
            \ForEach{$X_i$ in $S$}{
                sample event $Y$ uniformly from any chain in $C$ \;
                \uIf{$X_i$ is an ancestor of $Y$ in $G_c$}{
                    \texttt{$S$.add(counterfactual$_+$($X_i, Y$))}
                }\lElse{ \texttt{$S$.add(counterfactual$_-$($X_i, Y$))} }
            }
        }
        \texttt{$D$.add($S$)} \;
    }
	\caption{Pseudocode to generate synthetic relational data from causal graph $G_c$ and non-causal relation graph $G_n$. The helper-function \texttt{sample\_event\_chains} is described in \cref{alg:sample_event_chains}.}
	\label{alg:cap}
\end{algorithm}

\paragraph{Generating Causal Graphs.}

To generate a synthetic causal graph, we generate a directed acyclic graph with $n$ vertices and $r$ root vertices. Each vertex represents an event, and the root vertices are those that have no causes (i.e. they have no incoming edges). The algorithm to generate such a graph is shown in \cref{alg:gen_causal_graph}. The algorithm is fairly simple, but we take care not to create vertices that are descendants of all roots, since they will be causally connected to every root, and therefore, they would never be sampled in any event chain in \cref{alg:cap}. In addition, we require that every root has at least one child, in order to prevent generating trivial event chains that contain only a single event. In our experiments, we fix $n = 100$, and $r$ is sampled from $\text{Geometric}(0.64)$ conditioned on $r\in [3,6]$.

\begin{algorithm}
\footnotesize
\let\oldnl\nl
\newcommand{\nonl}{\renewcommand{\nl}{\let\nl\oldnl}}
\SetNlSty{}{\color{RedOrange}\sffamily}{}
\SetAlgoBlockMarkers{}{}
\SetKwProg{Fn}{function}{}{}
\SetKwIF{If}{ElseIf}{Else}{if}{ }{else if}{else }{}
\SetKw{Continue}{continue}
\SetKwFor{For}{for}{do}{end}
\SetKwProg{uForEach}{for each}{ do}{}
\SetKwProg{ForEach}{for each}{ do}{end}
\SetKwProg{Fn}{function}{}{}
\SetKwFor{RepTimes}{repeat}{times}{end}
\AlgoDisplayBlockMarkers\SetAlgoVlined
\SetAlCapNameFnt{\small}
\SetAlCapFnt{\small}
\SetNoFillComment
\DontPrintSemicolon
\SetInd{0.0em}{0.8em}
    \KwIn{number of vertices $n$, number of roots $r$}
    \KwOut{causal graph $G_c$}
    initialize $G_c$ as a graph with $n$ vertices and no edges \;
    let $(v_1,\hdots,v_n)$ be the vertices of $G_c$ \;
    \For{$i$ in $r+1,\hdots,n$}{
        $m \sim \text{Zipf}(3)$ \;
        $m \gets \min(i, m)$ \;
        sample $P$, a set of $m$ vertices from $\{v_1,\hdots,v_{i-1}\}$, uniformly without replacement \;
        \For{$p$ in $P$}{
            add edge $p\to v$ to $G_c$ \;
            \If{$v$ is a descendant of all roots $v_1,\hdots,v_r$}{
                remove edge $p\to v$ from $G_c$ \;
            }
        }
    }
    \tcc{make sure each root has $\ge$ 1 child}
    \For{$v_i$ in $\{v_1,\hdots,v_r\}$}{
        \If{$v_i$ has no child vertices}{
            $v \sim \text{Uniform}(v_{r+1},\hdots,v_n)$ \;
            add edge $v_i \to v$ to $G_c$ \;
        }
    }
    shuffle the vertices $(v_1,\hdots,v_n)$ \;
	\caption{Pseudocode for generating a synthetic causal graph.}
	\label{alg:gen_causal_graph}
\end{algorithm}

\paragraph{Generating Non-causal Relation Graphs.}

\cref{alg:gen_noncausal_graph} describes how we generate non-causal relations for the events in the causal graph. The output is a graph $G_n$ where each vertex represents a \emph{type} of event, and the function constructs a map $T$ from events in $G_c$ to event types in $G_n$. We chose simple semantics for $G_n$: If two events have the same type, they co-occur. An edge $a\to b$ in $G_n$ from event type $a$ to event type $b$ indicates that all events of type $a$ precede events of type $b$.

\begin{algorithm}
\footnotesize
\let\oldnl\nl
\newcommand{\nonl}{\renewcommand{\nl}{\let\nl\oldnl}}
\SetNlSty{}{\color{RedOrange}\sffamily}{}
\SetAlgoBlockMarkers{}{}
\SetKwProg{Fn}{function}{}{}
\SetKwIF{If}{ElseIf}{Else}{if}{ }{else if}{else }{}
\SetKw{Continue}{continue}
\SetKwFor{For}{for}{do}{end}
\SetKwProg{uForEach}{for each}{ do}{}
\SetKwProg{ForEach}{for each}{ do}{end}
\SetKwProg{Fn}{function}{}{}
\SetKwFor{RepTimes}{repeat}{times}{end}
\AlgoDisplayBlockMarkers\SetAlgoVlined
\SetAlCapNameFnt{\small}
\SetAlCapFnt{\small}
\SetNoFillComment
\DontPrintSemicolon
\SetInd{0.0em}{0.8em}
    \KwIn{causal graph $G_c$}
    \KwOut{non-causal relation graph $G_n$}
    let $(t_1,\hdots,t_k)$ be (an initially empty) ordered list of event types \;
    let $T$ be an initially empty map from events in $G_c$ to event types $\{t_1,\hdots,t_k\}$ \;
    \ForEach{event $v$ in $G_c$}{
    \tcc{\fontsize{7.6}{9}\selectfont assign an event type to each event in $G_c$}
        compute $\alpha = \max\{i : \text{there is an ancestor }a\text{ of }v\text{ such that }T(a)=t_i\}$ \;
        compute $\beta = \min\{i : \text{there is a descendant }d\text{ of }v\text{ such that }T(d)=t_i\}$ \hspace{-1em} \;
        \uIf{$\alpha < \beta$}{
            $w \sim \text{Uniform}(t_{\alpha+1},\hdots,t_{\beta-1})$ \;
        }\Else{
            create new event type $w$ and insert it into the list of event types at index $\alpha + 1$ \;
        }
        set $T(v) \gets w$ \;
    }
    let $(t_1,\hdots,t_k)$ be the vertices of $G_n$ \;
    \tcc{\fontsize{8.5}{9}\selectfont add temporal edges between event types}
    \ForEach{event $v$ in $G_c$}{
        \ForEach{child vertex $c$ of $v$}{
            add edge $T(p) \to T(c)$ to $G_n$ \;
        }
    }
	\caption{Pseudocode for generating a synthetic non-causal relation graph.}
	\label{alg:gen_noncausal_graph}
\end{algorithm}

\paragraph{Sampling Event Chains.}

\cref{alg:sample_event_chains} describes the helper function used in \cref{alg:cap} which samples a handful of event chains, where each chain is causally-independent of the other event chains. In this helper function, each event chain starts at a root node in $G_c$, since root nodes are by definition causally-independent of each other. We sample the length of each chain to be uniform so that vertices near roots are not over-represented in the sample of event chains (and vertices further from the roots are not under-represented). This helps to facilitate more uniform coverage of all vertices in $G_c$ by the generated data.

\begin{algorithm}
\footnotesize
\let\oldnl\nl
\newcommand{\nonl}{\renewcommand{\nl}{\let\nl\oldnl}}
\SetNlSty{}{\color{RedOrange}\sffamily}{}
\SetAlgoBlockMarkers{}{}
\SetKwProg{Fn}{function}{}{}
\SetKwIF{If}{ElseIf}{Else}{if}{ }{else if}{else }{}
\SetKw{Continue}{continue}
\SetKwFor{For}{for}{do}{end}
\SetKwProg{uForEach}{for each}{ do}{}
\SetKwProg{ForEach}{for each}{ do}{end}
\SetKwProg{Fn}{function}{}{}
\SetKwFor{RepTimes}{repeat}{times}{end}
\AlgoDisplayBlockMarkers\SetAlgoVlined
\SetAlCapNameFnt{\small}
\SetAlCapFnt{\small}
\SetNoFillComment
\DontPrintSemicolon
\SetInd{0.0em}{0.8em}
    \KwIn{causal graph $G_c$}
    \KwOut{set of event chains $C$}
    initialize $C \gets \{\}$ \;
    $n \sim 1 + \text{Geometric}(0.25)$ \;
    sample $R$, a set of $n$ root vertices from $G_c$ (with no incoming edges), uniformly without replacement \;
    \tcc{for each root, sample a chain}
    \ForEach{$r$ in $R$}{
        compute $D_r$, the set of descendant vertices of $r$ \;
        \tcc{sample the length of this chain}
        $m \sim \text{Uniform}(1, \hdots, \max_{v\in D_r} \text{distance}(r, v))$ \;
        compute $S_{r,m} = \{v \in D_r : \text{distance}(r,v) = m$ $\text{ and } v \text{ is not a descendant of } R \setminus \{r\} \}$ \;
        \While{$S_{r,m}$ is empty}{
            $m \gets m - 1$ \;
            recompute $S_{r,m}$ as above \;
        }
        \tcc{sample the endpoint of the chain}
        $e \sim \text{Uniform}(S_{r,m})$ \;
        $C\texttt{.add(}\text{set of all vertices on path from } r \text{ to } e\texttt{)}$ \;
    }
    \tcc{mark some chains as `non-occurring'}
    $k \sim \text{Binomial}(n - 1, 0.2)$ \;
    remove $k$ event chains from $C$, uniformly at random \;
	\caption{Pseudocode for the helper-function \texttt{sample\_event\_chains}, which, given a causal graph $G_c$, returns a number of event chains, where each chain is causally-independent of the other chains.}
	\label{alg:sample_event_chains}
\end{algorithm}

\paragraph{Generating Scenarios.} \cref{alg:cap} gives the data generation algorithm for generating the scenarios. In each step, when we sample $S$, a set of $n$ events from the $\texttt{event\_chain}$ we sample uniformly randomly without replacement. This ensures that scenarios contain information about a diverse set of events.

We also include an example from our generated dataset, where the scenario contains all three relations in \cref{tab:example}.

\begin{table*}[t]
\begin{small}
    \centering
    \begin{tabularx}{\linewidth}{lX}
    \toprule
        All Relations &  event84 happened. event76 happened. event76 and event84 took place in the same location. if event76 did not happen, and event84 has no other causes, would event84 happen? yes. if event76 has no other causes, and event84 did not occur, would event76 still happen? no. event5 happened. event3 happened. event96 happened. event3 happened after event84. event5 happened before event3. the location of event96 is not identical to that of event76. if event3 did not happen, and event5 has no other causes, would event5 happen? yes. \\
        Temporal Relations  & event67 occurred prior to event71. event40 happened before event28. event7 preceded event28. event71 happened after event95. \\
        Spatial Relations & the location of event96 is not identical to that of event4. event4 and event96 did not take place in the same location. \\
        Counterfactuals & if event33 did not occur, and event84 has no other causes, would event84 still happen? yes. if event84 has no other causes, and event58 did not occur, would event84 still happen? yes. if event58 has only one cause, and hypothetically event84 did not happen, would event58 still occur? no. if event3 has only one cause, and event48 did not happen, would event3 happen? yes.\\
        \bottomrule
    \end{tabularx}
    \caption{Examples of the scenarios from our generated dataset. The first examples contains all types of relations, whereas the others include one type of relation only.}
    \label{tab:example}
\end{small}
\end{table*}

\subsection{Experiment Details}
\label{app:exp_details}

We used \textsc{Llama2} models through HuggingFace's transformer library \cite{Wolf2019HuggingFacesTS}. All models were finetuned with LoRA (applied to query and key projection matrices), with rank $= 16$, $\alpha = 16$ and dropout $=0.05$. All models were finetuned with a learning rate of $5e-4$ using AdamW optimizer\cite{DBLP:journals/corr/KingmaB14, Loshchilov2017DecoupledWD} with a batch size of 8. The models finetuned on 36k scenarios were trained for 10k steps whereas the models trained with 4.5k scenarios (500 scenarios used for validation, as in \cref{app:freq}) were trained for 6k steps --- we generally observed that models converged around this point.

\subsection{Additional Results: Position Bias}
\label{app:add_details}

\paragraph{Temporal Relations.}
\cref{ssec:position_order} showed that, in the presence of strong position bias, the model assigned high probability to $t_i(X \rightarrow Y)$ where the relative position matches that during finetuning. This still leaves open the possibility that the model is assigning a higher probability to the template for \emph{correct} causal relation where the position matches. e.g. from `$X$ preceded $Y$', the model could assign probabilities in the following order --- `$X$ can cause $Y$' $>$ `$X$ can be caused by $Y$' $>$ `$Y$ can be caused by $X$' $>$ `$Y$ can cause $X$'. In such a situation, if the order is randomized during evaluation the model can still infer causal relations from temporal relations.

In this experiment, we find that models finetuned on temporal relations with relative position $(X, Y)$ infer $X \rightarrow Y$ from $\texttt{temporal}(X, Y)$ 23.14\% of the times. Since random chance is 33.3\%, we see that models finetuned on position bias indeed are not able to make any consistent deduction beyond matching relative position during finetuning and evaluation.

\begin{table}[t]
\begin{small}
    \centering
    \begin{tabular}{cccc}
    \toprule
     \multirow{2}{*}{Data} & {Rel. position}   & \multicolumn{2}{c}{Rel. position in eval}  \\
       &  in train  &  $(X, Y)$ & $(Y, X)$ \\
    \midrule

    \multirow{2}{*}{causal $X\to Y$} & $(X, Y)$ & \cellcolor{lightapricot}92.59\% & 1.85\% \\
     &   $(Y, X)$ & 0\% & \cellcolor{lightapricot}100\% \\

    \midrule
    \multirow{2}{*}{unrelated $X, Y$} & $(X, Y)$ & \cellcolor{lightapricot}98.14\% & 0.92\% \\
    & $(Y, X)$ & 0\% & \cellcolor{lightapricot}100\% \\
    
    \bottomrule
    \end{tabular}
    \caption{Accuracy of models finetuned on temporal relations with different relative event positions. Models infer the causal relation only when the relative position \colorbox{lightapricot}{matches} during finetuning and evaluation. 
    }
    \label{tab:app_position_eval}
\end{small}
\end{table}

To further show that models are only relying on the relative position of events instead of reasoning about their causal relation, we evaluate models using different relations with the \emph{same} relative position.
Specifically, we randomly sample three relations between $X$ and $Y$ which have no connection to the causal relation and verbalize them using the $(X, Y)$ relative order e.g. instead of the verbalization `$X$ causes $Y$', we will use `$X$ is related to $Y$' (details in \cref{app:templates}).
We observe a similar result in the last two rows in \cref{tab:app_position_eval}---models only make correct predictions when the event order during training matches that during test.

\begin{table}[t]
\begin{small}
    \centering
    \begin{tabular}{cccc}
    \toprule
     & {Rel. position}   & \multicolumn{2}{c}{Rel. position - eval}  \\
       &  during train  &  $(X, Y)$ & $(Y, X)$ \\
    \midrule

    Accuracy & $(X, Y)$ & \cellcolor{lightapricot}90.5\%/3.0\% & 6.5\%/3.5\% \\
    
    \bottomrule
    \end{tabular}
    \caption{Models finetuned on spatial relations with fixed relative position, and we report \% of cases model infer $X \rightarrow Y$ / \% of cases model infers $X \nleftrightarrow Y$. Models infer the causal relation only when the relative position \colorbox{lightapricot}{matches} during finetuning and evaluation.
    }
    \label{tab:position_bias_spatial}
\end{small}
\end{table}

\paragraph{Spatial Relations.}
Here, we demonstrate that we also observe the position bias for spatial relations. To show this we first create a dataset with fixed relative position. Specifically, we generate a dataset $D_{\texttt{spatial}, (X, Y)}$ consisting of positive and negative spatial relations from the sets $T(\texttt{spatial}_+, (X, Y))$ and $T(\texttt{spatial}_-, (X, Y))$ respectively.
We then finetune $\textsc{Llama2-7B}$ on this data and evaluate the model on $D_{\text{unrelated}X-Y}$. We use two different sets of templates to evaluate the model: $T(X \rightarrow Y, (X, Y))$ (e.g. `$X$ causes $Y$') or using templates from  $T(X \rightarrow Y, (Y, X))$ (e.g. `$Y$ is caused by $X$'). In both cases, to score the relation $X \nleftrightarrow Y$ we use $T(X \nleftrightarrow Y, (X, Y)+(Y,X))$.

\cref{tab:position_bias_spatial} shows the percentage of examples in which the model predicted either $X \rightarrow Y$ or $X \nleftrightarrow Y$ (which is the correct option).
Firstly, we observe that in both cases, the model rarely selects the correct option $X \nleftrightarrow Y$.
Similar to the position bias in temporal relations, the model selects either $X \rightarrow Y$ depending on if the position matches. This shows that position bias also exists for spatial relations.
We also evaluate the model using templates which have randomized relative position for each option. Specifically, we use templates from the sets $T(r, (X, Y) + (Y, X))$ where $r \in \{X \rightarrow Y, Y \rightarrow X, X \nleftrightarrow Y\}$. We find that model selects the correct option ($X \nleftrightarrow Y$), 68\% of the time. This is in contrast to the position bias in temporal relations, where the performance was close to random chance. Nevertheless, the model still performs worse than if the position was randomized in the finetuning data (84.5\%, \cref{tab:randomized_pos})

In summary, we find that the position bias also holds true for spatial relations, albeit to a lesser extent than that for temporal relations.

\subsection{Position heuristic is supported in the pretraining data}
\label{app:pile}

\cref{ssec:position_order} demonstrated that LLMs fail to infer causal relations if the finetuning data supports the position heuristic.
We hypothesize that this phenomenon occurs since the position heuristic is supported in the pretraining data --- if cause is often mentioned before effect in the text, then LLMs can use relative position as a heuristic for the language modeling task. E.g. for the causal relation `smoking causes cancer', we hypothesize that `smoking' usually occurs before `cancer' if they co-occur within a window.
Thus a LLM trained on such data can do well even if it only uses the heuristic of relative position to predict the next word and ignore the relation between the two events.

To test if this holds true in the pretraining data, for a given causal relation $X \rightarrow Y$, we count the number of times $X$ occurs before or after $Y$ in a context window. We expect that if the heuristic is supported in the pretraining data, then $X$ should mostly occur before $Y$ when they co-occur in a context window.

We first create a set of 40 commonly-queried causal relations (e.g. smoking causes cancer, bacteria causes infections, etc.) based on the edges from the CauseNet dataset \cite{heindorf2020causenet}, the Tubingen dataset \cite{Mooij2014DistinguishingCF} as well as some candidates from GPT-4.
Then for each of the causal relations $X \rightarrow Y$, we count the number of documents of the PILE\footnote{The pretraining dataset for \textsc{Llama2-7B} is not available, so we use PILE and assume that relative positions would be similar.}  corpus \cite{Gao2020ThePA} in which either $X$ occurs before $Y$ or $Y$ occurs before $X$ within a window of 50 characters of the first mention of $X$ and $Y$ in the document. 
We filter to keep only those edges where the events co-occur within the context window at least 100 times. See \cref{app:pile_relations} for details.

Across all causal relations, we find that whenever $X$, $Y$ co-occur within the context window, $60.77\%$ of the times $X$ occurs before $Y$. Overall, we observe that the data supports the heuristic in a majority (> 50\%) of the examples.

\subsection{Additional Results: Frequency vs Position Bias}
\label{app:freq}

We also observe an interesting trend where models exhibit a stronger position bias for relations that are more frequent in the finetuning data.
To show this, we first create a smaller dataset by sampling 5k examples from $D_{\texttt{temporal}, (X, Y)}$ --- 4.5k for finetuning, 500 for evaluation --- and finetune for fewer steps.
We split the test set $D_{X \rightarrow Y}$ into 10 equal sized buckets based on the frequency of the corresponding temporal relation, $\texttt{temporal}(X, Y)$, in $D_{\texttt{temporal}, (X, Y)}$.

\begin{figure}
    \centering
    \includegraphics[scale=0.56]{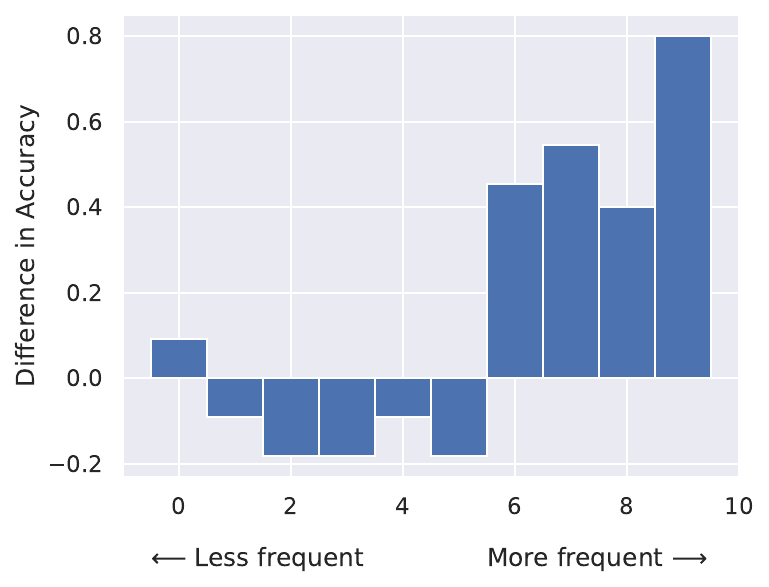}
    \caption{Difference in accuracy on the test sets with matched and unmatched event orders as a function of the frequency of the relation in the data. LLMs suffer from the position bias on high frequency events.
    }
    \label{fig:frequency}
\end{figure}

\cref{fig:frequency} shows the result where the X-axis is the frequency buckets, and Y-axis is the difference in accuracy between the test set with matched and unmatched X-Y orders. We observe that  high frequency relations are correlated with a larger gap. 

\begin{table}[t]
\begin{small}
    \centering
    \begin{tabular}{cccc}
    \toprule
     & {Rel. position}   & \multicolumn{2}{c}{Rel. position - eval}  \\
       &  during train  &  $(X, Y)$ & $(Y, X)$ \\
    \midrule

    \multirow{2}{*}{Three-way eval} & $(X, Y)$ & 52.77\% & 35.18\% \\
    &   $(Y, X)$ & 3.70\% & 94.44\% \\
    \bottomrule
    \end{tabular}
    \caption{Models finetuned on 5k scenarios with temporal relations with different relative positions. We only observe the position effect in one direction (when finetuned on $(Y, X)$) but not the other.}
    \label{tab:position_eval_5k}
\end{small}
\end{table}

We also report the absolute accuracy when the model is trained on the smaller finetuning dataset with 4.5k scenarios. As shown before, in this case we observed the position bias for high frequency relations. In \cref{tab:position_eval_5k}, we report the avg accuracy of models inferring $X \rightarrow Y$ for both relative positions. 
We observe a stronger position effect in one direction (when trained with relative position $(Y, X)$) but not as much in the other direction. Note that the model performance when trained with relative position $(X, Y)$ is not much better than chance and is also sensitive to the relative position.

\subsection{Additional Results: Alternate evaluation of $X \rightarrow Y$}
\label{app:alternate_eval}

In \cref{ssec:eval} to evaluate models, we first compute the probabilities of the following five relations under the language model: $X \rightarrow Y$, $Y \rightarrow X$, $X \not\rightarrow Y$, $Y \not\rightarrow X$, and $X \nleftrightarrow Y$. To test if models have inferred the causal relation $X \rightarrow Y$, we compare the probabilities of the following three events which are exhaustive (i.e. their true probabilities sum to 1) and disjoint: $X \rightarrow Y$, $Y \rightarrow X$, and $X \nleftrightarrow Y$.

An alternative set of events which are also exhaustive and disjoint are: $X \rightarrow Y$, and $X \not\rightarrow Y$. In this section, we demonstrate that our conclusion of whether models infer $X \rightarrow Y$ remains consistent even if we use these two events as the set of events to compare.

To show this, we re-evaluate two models: \textsc{Llama2-7B} finetuned on $D_{\texttt{temporal}}$, and $D_{\texttt{counterfactual}}$ respectively. We then evaluate these models on $D_{\text{causal}X-Y}$ to test if they infer presence of causal relations from either temporal relations or positive counterfactuals.

\begin{table}[t]
\begin{small}
    \centering
    \begin{tabular}{cc}
       \toprule
       & $D_{\text{causal}X-Y}$\\
       \midrule
       $\texttt{temporal}(X, Y) \implies X \rightarrow Y$  & 71.29\%\\
       $\texttt{counterfactual}_+(X, Y) \implies X \rightarrow Y$ & 54.62\%\\
       \bottomrule
    \end{tabular}
    \caption{Alternative Evaluation: Using a different set of exhaustive and disjoint events does not change our conclusions --- model suffer from post hoc fallacy, and they cannot infer presence of causal relation from counterfactual.
    }
    \label{tab:alternative_eval}
\end{small}
\end{table}

\cref{tab:alternative_eval} shows the percentage of examples where model predicts the causal relation $X \rightarrow Y$. First, we observe that models infer causal relations from temporal relation --- i.e. $\texttt{temporal(X, Y)} \implies X \rightarrow Y$. Therefore, similar to our previous findings where models suffer from post hoc fallacy (\cref{sec:post_hoc_fallacy}), changing how we evaluate the presence of causal relation does not affect our results. Similarly, we observe that models cannot infer presence of causal relations from counterfactuals much better than random chance (50\%). This is consistent with our finding from \cref{sec:inference_results}, where we showed that the model cannot infer causal relations from positive counterfactuals.

\subsection{Templates for Relations}
\label{app:templates}

In this section, we list the templates we use for each of the three relations: temporal relations, spatial relations, and counterfactuals. Additionally, we also describe the templates we used for causal relations (both presence and absence of causal relations). Each template is separated by `;'.

\begin{enumerate}
    \item $T(\texttt{temporal}(X, Y), (X, Y))$: $X$ preceded $Y$; $X$ happened before $Y$; $X$ occurred prior to $Y$; $X$ took place before $Y$; $X$ happened then $Y$ happened
    \item $T(\texttt{temporal}(X, Y), (Y, X))$: $Y$ followed $X$; $Y$ happened after $X$; $Y$ occurred later than $X$; $Y$ took place after $X$; $Y$ happened later than $X$
    \item $T(\texttt{temporal}(X, Y), \texttt{random})$: $X$ preceded $Y$; $Y$ followed $X$; $X$ occurred prior to $Y$; $Y$ happened after $X$; $Y$ occurred later than $X$; $X$ happened before $Y$
    \item $T(\texttt{spatial}_+(X, Y), \texttt{random})$: $X$ and $Y$ took place in the same location; the location of $X$ is identical to that of $Y$; $X$ and $Y$ happened in the same place; $Y$ and $X$ took place in the same location; the location of $Y$ is identical to that of $X$; $Y$ and $X$ happened in the same place
    \item $T(\texttt{spatial}_-(X, Y), \texttt{random})$: $X$ and $Y$ did not take place in the same location; the location of $X$ is not identical to that of $Y$; $X$ and $Y$ did not happen in the same place; $Y$ and $X$ did not take place in the same location; the location of $Y$ is not identical to that of $X$; $Y$ and $X$ did not happen in the same place
    \item $T(\texttt{counterfactual}_+(X, Y), \texttt{random})$: if $X$ did not happen, and $Y$ has no other causes, would $X$ happen? no; if $Y$ has only cause, and $X$ did not happen, would $Y$ happen? no; if $X$ did not occur, and $Y$ has no other causes, would $Y$ still happen? no; if $Y$ has no other causes, and $X$ did not occur, would $Y$ still happen? no; if hypothetically $X$ did not happen, and $Y$ has only cause, would $Y$ still occur? no; if $Y$ has only cause, and hypothetically $X$ did not happen, would $X$ still occur? no; 
    \item $T(\texttt{counterfactual}_-(X, Y), \texttt{random})$ if $X$ did not happen, and $Y$ has no other causes, would $X$ happen? yes; if $Y$ has only cause, and $X$ did not happen, would $Y$ happen? yes; if $X$ did not occur, and $Y$ has no other causes, would $Y$ still happen? yes; if $Y$ has no other causes, and $X$ did not occur, would $Y$ still happen? yes; if hypothetically $X$ did not happen, and $Y$ has only cause, would $Y$ still occur? yes; if $Y$ has only cause, and hypothetically $X$ did not happen, would $X$ still occur? yes; 
    \item $T(X \rightarrow Y, \texttt{random})$: $X$ can cause $Y$; $Y$ can be caused by $X$; $X$ causally affects $Y$; $X$ can lead to $Y$; $Y$ is causally affected by $X$; $Y$ is caused by $X$
    \item $T(X \not\rightarrow Y, \texttt{random})$: $X$ cannot cause $Y$; $Y$ cannot be caused by $X$; $X$ does not causally affects $Y$; $X$ cannot lead to $Y$; $Y$ is not causally affected by $X$; $Y$ is not caused by $X$
    \item $T(X \nleftrightarrow Y, \texttt{random})$: `there is no causal relation between $X$ and $Y$', `there is no causal relation between $Y$ and $X$', `there is no dependency between $X$ and $Y$', `there is no dependency between $Y$ and $X$', `there is no causal link between $X$ and $Y$', `there is no causal link between $Y$ and $X$',
    `$X$ neither causes nor is caused by $Y$',
    `$Y$ neither causes nor is caused by $X$',
    `there is no cause-and-effect relationship between $X$ and $Y$', `there is no cause-and-effect relationship between $Y$ and $X$', `there is no causal association linking $X$ and $Y$', `there is no causal association linking $Y$ and $X$'
\end{enumerate}

\subsection{Position Heuristic in PILE}
\label{app:pile_relations}

For searching through the pretraining data, we used the PILE corpus since it's freely available and has been used in recent models e.g. Pythia models \cite{Biderman2023PythiaAS}.
Here, we list the 40 causal relations we used to search over the PILE corpus. We set the parameter $w$ to be 50 characters i.e. the events are said to co-occur if they occur within 50 characters of each other. We filter to keep only those edges where the events co-occur enough times in the pretraining data (we set it to 100) --- this is done to ensure that results are not affected by causal relations where the events do not frequently co-occur. 

\begin{verbatim}
[('bacteria', 'infections'),
 ('hiv', 'aids'),
 ('cancer', 'death'),
 ('smoking', 'lung cancer'),
 ('altitude', 'temperature'),
 ('age', 'height'),
 ('sun exposure', 'aging'),
 ('sugar', 'tooth decay'),
 ('drugs', 'organ damage'),
 ('salt', 'high blood pressure'),
 ('screens', 'eye strain'),
 ('lack of sleep', 'impaired cognition'),
 ('pollution', 'lung harm'),
 ('noise', 'hearing loss'),
 ('genetics', 'height'),
 ('dehydration', 'fatigue'),
 ('sugar', 'diabetes'),
 ('stress', 'headache'),
 ('poor nutrition', 'fatigue'),
 ('sedentary habits', 'obesity'),
 ('education', 'income'),
 ('physical activity', 'health'),
 ('parental involvement', 'child development'),
 ('nutrition', 'longevity'),
 ('financial stress', 'mental health'),
 ('pollution', 'health problems'),
 ('stress', 'immune function'),
 ('education', 'political participation'),
 ('drugs', 'crime rate'),
 ('deforestation', 'climate change'),
 ('fossil fuels', 'climate change'),
 ('greenhouse gases', 'climate change'),
 ('accident', 'death'),
 ('stroke', 'death'),
 ('diabetes', 'death'),
 ('migraine', 'headache'),
 ('smoking', 'house fires'),
 ('infidelity', 'divorce'),
 ('poverty', 'homelessness'),
 ('drunk driving', 'accident')]
\end{verbatim}

\end{document}